# CarDS-Plus ECG Platform
## Development and Feasibility Evaluation of a Multiplatform Artificial Intelligence Toolkit for Portable and Wearable Device Electrocardiograms

Sumukh Vasisht Shankar, MS [1,2], Evangelos K Oikonomou, MD DPhil [1,2], Rohan Khera, MD MS [1,2,3,4]

1. Cardiovascular Data Science (CarDS) Lab, Yale School of Medicine, New Haven, CT
2. Section of Cardiovascular Medicine, Department of Internal Medicine, Yale School of Medicine, New Haven, CT
3. Center for Outcomes Research and Evaluation (CORE), Yale New Haven Hospital, New Haven, CT, USA
4. Section of Health Informatics, Department of Biostatistics, Yale School of Public Health, New Haven, CT, USA

**Corresponding author:**
Rohan Khera, MD MS
195 Church Street, 6th Floor, New Haven, CT 06510
rohan.khera@yale.edu

**Keywords:** Artificial Intelligence; Wearable Devices; Electrocardiograms;

**Abstract**

In the rapidly evolving landscape of modern healthcare, the integration of wearable and portable technology provides a unique opportunity for personalized health monitoring in the community. Devices like the Apple Watch, FitBit, and AliveCor KardiaMobile have revolutionized the acquisition and processing of intricate health data streams that were previously accessible only through devices only available to healthcare providers. Amidst the variety of data collected by these gadgets, single-lead electrocardiogram (ECG) recordings have emerged as a crucial source of information for monitoring cardiovascular health. Notably, there has been significant advances in artificial intelligence capable of interpreting these 1-lead ECGs, facilitating clinical diagnosis as well as the detection of rare cardiac disorders. This design study describes the development of an innovative multi-platform system aimed at the rapid deployment of AI-based ECG solutions for clinical investigation and care delivery. The study examines various design considerations, aligning them with specific applications, and develops data flows to maximize efficiency for research and clinical use. This process encompasses the reception of single-lead ECGs from diverse wearable devices, channeling this data into a centralized data lake, and facilitating real-time inference through AI models for ECG interpretation. An evaluation of the platform demonstrates a mean duration from acquisition to reporting of results of 33.0 to 35.7 seconds, after a standard 30 second acquisition, allowing the complete process to be completed in 63.0 to 65.7 seconds. There were no substantial differences in acquisition to reporting across two commercially available devices (Apple Watch and KardiaMobile). These results demonstrate the succcessful translation of design principles into a fully integrated and efficient strategy for leveraging 1-lead ECGs across platforms and interpretation by AI-ECG algorithms. Such a platform is critical to translating AI discoveries for wearable and portable ECG devices to clinical impact through rapid deployment.

## 1  Background

In the dynamic landscape of modern healthcare, the integration of wearable and portable technology has heralded a revolution in personalized health monitoring. Wearable health devices, such as the Apple Watch and Fitbit, and the portable AliveCor KardiaMobile offer access to complex health data streams that previously required large and complex devices for acquisition and processing. Among the various kinds of data collected by these devices, single-lead electrocardiogram (ECG) recordings stand out as a vital source of information for monitoring cardiovascular health. There has been a major development of artificial intelligence with capacity to interpret these 1-lead ECGs for both clinical diagnosis, as well as the detection of rare cardiac disorders (Dhingra **andothers** 2023).

Recognizing the significance of these data and the increasing adoption of wearable and portable devices, we sought to develop and evaluate a centralized approach that is capable to receive single-lead ECGs across various wearable devices, and enable real-time inference from advance artificial intelligence (AI) models developed for ECG interpretation.

## 1.1 Wearable ECG Monitoring

The use of wearable devices for ECG monitoring enables access to an important modality for informatoin on cardiovascular health. Devices like the Apple Watch, KardiaMobile, and Fitbit are now equipped with advanced sensors capable of recording single-lead ECGs. These ECGs capture the electrical activity of the heart, and with applications of AI aloowing evlauations of cardiac structure and from electrocardiographic data (Attia, Noseworthy **andothers** 2019, Attia, Kapa **andothers** 2019, Sangha **andothers** 2022, Neri **andothers** 2023, Oikonomou **and** Khera 2023). While these devices empower individuals to take charge of their health, harnessing the data consistently and translating it into actionable insights remains challenging.

## 1.2 Need for a Unified ECG Dashboard

The CarDS-Plus ECG Dashboard is a platform designed to integrate ECGs from various wearable devices. Patients, physicians, and healthcare providers increasingly require a unified solution to consolidate data for comprehensive assessment of health, especially with AI solutions allowing inference on multiple axes of health and outcomes.

In this design study, we describe the development of an innovative multi-platform for rapid cycle deployment of AI ECG solutions for clinical investigation and care delivery. We identify design considerations in light of the applications and define the data flow and security from receiving single-lead ECGs across various wearable devices in a centralized repository and real-time inference from advanced AI models developed for ECG interpretation.

# 2 Rationale and Guiding Principles

Cardiovascular disease remains a leading cause of mortality globally, necessitating innovative approaches to enhance diagnostic accuracy and expedite treatment decisions. Traditional ECG analysis methods, while effective, require specialized expertise and are limited to gross abnormalities that are detectable by the human observer. Integrating AI into portable ECG devices addresses these challenges, allowing for rapid and reliable interpretation of ECG data as well access to AI-guided analytics, even in resource-constrained environments. Moreover, the combination of smartphones and wearable devices provides an opportunity to create a multiplatform toolkit that can be seamlessly integrated into various healthcare settings, from remote clinics to emergency rooms.

## 2.1 Applications

- **Early Detection of Cardiac Abnormalities**: ECG models can detect subtle patterns and irregularities in ECG data that may not be apparent to human observers. Deploying these models at scale can screen a large number of individuals efficiently, identifying potential cardiac abnormalities. Early detection is crucial for initiating timely interventions and preventing serious cardiac events.

- **Remote Monitoring and Telehealth**: Scalable ECG models enable remote monitoring of patients' heart health. This is especially valuable for individuals with chronic conditions or those who live in remote areas with limited access to healthcare facilities. Through wearable devices and mobile apps, patients can record ECG data and have it analyzed by the models, providing real-time insights to both patients and healthcare providers without the need for in-person visits.

- **Standardization of Care**: ECG models can provide consistent and standardized interpretations of ECG data. This consistency is essential, especially in healthcare systems with varying levels of expertise among clinicians. Standardized interpretations help ensure that patients receive consistent care and reduce the likelihood of misdiagnoses or missed abnormalities.

- **Research and Development**: Scalable ECG models contribute to the advancement of medical knowledge. Researchers can analyze anonymized, aggregated ECG data from diverse sources to identify new patterns, risk factors, and treatment strategies for cardiac conditions. This can lead to the development of more effective therapies and preventive measures.

- **Patient Empowerment**: Scalable ECG models can also be integrated into consumer-facing applications and wearable devices, allowing individuals to monitor their own heart health proactively. These models can provide users with insights, alerts, and recommendations based on their ECG data, empowering them to take control of their health and seek medical attention when needed.

## 2.2 Design Considerations

A few elements critical to design this unified platform to collect and analyze ECG data from different devices include:

- **Deployment at Scale**: Deployment of machine learning (ML) models at scale is crucial because it enables the practical application of AI technologies in real-world scenarios. This scalability is essential for our use case, where large number of ECG data samples are generated daily. Scaling ML models ensures that valuable insights are leveraged, operational efficiency is improved, and informed decisions are made in time.

- **Multi-Device Compatibility**: Our ECG Dashboard is compatible with a range of popular wearable devices, including the Apple Watch, Kardia, and Fitbit. Users can effortlessly connect and sync data from their chosen devices, allowing for a holistic view of their heart's performance over time. As far as we are aware, there is no single platform that can collect ECG data from all three devices and run predictive models on them.

- **Real-Time Data Visualization**: The dashboard provides real-time visualization of ECG data, enabling users to monitor their heart rhythms as they happen.

This capability empowers individuals to detect irregularities promptly and take immediate action when necessary.

- **Historical Data Analysis**: In addition to real-time data, our platform stores historical ECG recordings in compliance to the Health Insurance Portability and Accountability Act, allowing users to track their heart health progress over days, weeks, or months. This longitudinal perspective is invaluable for identifying trends and patterns in cardiovascular health.

- **Predictive Analysis**: One of the most important features of our ECG Dashboard is its predictive analytics engine. By leveraging advanced machine learning algorithms and artificial intelligence, the dashboard can analyze ECG data and provide predictions for various heart conditions within seconds. These predictions encompass several conditions, ranging from the real-time monitoring of rhythm disorders, to dynamic prediction of left ventricular function, development of valvular abnormalities and response to novel disease-modifying therapies. This predictive capability empowers users to proactively manage their heart health and seek medical attention when necessary.

In an era where data-driven approaches play an important role in healthcare decision-making, our ECG Dashboard bridges the gap between wearable technology and actionable insights. By bringing together ECG data from multiple devices, offering real-time monitoring, and predictive analytics, our platform empowers individuals to take control of their heart health. It has the potential to support clinicians in making informed decisions, and it contributes to the body of knowledge in cardiac health research.

## 3 Methods

The development of the deep learning tools were reviewed by the Yale Institutional Review Boards (IRB), which approved the study protocol and waived the need for informed consent as the study represents secondary analysis of existing data (Yale IRB ID #2000029973). No patient data were used in the evaluation of the model, with only deidentified recordings used in the time trials, and were not human subject research.

The objective of this design study was to create a seamless and accurate process for collecting ECG data from wearable and portable devices, analyzing the data, making predictions about heart conditions, and feeding the results into a data pipeline for visualization on a dashboard. Below is a detailed step-by-step description of the proposed method:

- **Data Collection**:
    - **Data Sources**: ECG data are collected from multiple wearable devices, including the Apple Watch, Kardia and Fitbit. The process of collecting data from each of these devices can vary substantially.

- **Apple Watch**: Out of the three devices our platform is compatible with, collecting ECG data from Apple watches is a lengthy process. There exist no single API that can fetch health data from a user's Apple Health Store. One way to mitigate this problem is by developing an iOS application that, with the permission of the user to enable access to Apple Health Store, can collect the data, and transmit it to our database where predictive analysis will be run before displaying it on the dashboard, as well as on the iOS app itself. Figure 3 shows a screenshot of our prototype application, and its icon. The user will have the control of choosing which ECG to submit for predictive analysis and displaying on the dashboard.
- **Fitbit**: Collecting ECG data from FitBit is a relatively more straightforward process compared to Apple Watch. Google has its own FitBit API (Application Programming Interface) that can be used to collect data from the cloud. It however requires the permission of the FitBit user to do so. Once the permission is granted, data can seamlessly be collected from Google Cloud by using the FitBit ID of the user.
- **Kardia**: Collecting ECG data from Kardia is similar to that of Fitbit, but does not require any explicit permission as far as the APIs are concerned. The device will be registered to be used with our dashboard and data from only those devices can be collected. Kardia has its own API which can be used to query the required data.

- **Data Transmission**: Data collected are transmitted to a secure cloud-based server or platform (AWS S3 bucket in compliance to HIPAA standards) in real-time.
- **Data Storage**: ECG data is stored in a secure, scalable, and compliant data storage system. A combination of relational databases and cloud-based storage solutions, such as Amazon S3 and EC2 instances, are used to store raw data and metadata.

- **Data preprocessing**:
  - **Data Cleaning**: Raw ECG data often contains noise and artifacts. Data cleaning processes, such as filtering and noise reduction, are applied to enhance data quality. This is done according to established data preprocessing methods, with the addition of noise adjustment tools developed by our team. (Khunte **andothers** 2023)
  - **Data Integration**: Data from different wearable devices may have varying formats and standards. Data integration techniques are employed to standardize and unify the data for consistent analysis.

- **Predictive Analysis**:
  - **Machine Learning Models**: Machine learning algorithms are trained on historical ECG data with labeled outcomes to predict heart conditions. Models may include binary classifiers (e.g., normal vs. abnormal) or multi-class classifiers for specific conditions.

- **Real-Time prediction**: For real-time predictions, streaming data from wearable devices is continuously analyzed by the trained models. Predictions are generated as soon as new data becomes available and displayed on the dashboard within a few seconds.

- **Data Pipeline Integration**:
  - **API integration**: Predictions and analysis results are exposed through APIs (Application Programming Interfaces) for seamless integration into the data pipeline.
  - **Data Transformation**: Predictions and results are formatted into a compatible data structure for ingestion into the data pipeline. This may involve converting results into JSON or XML formats.
  - **Data Pipeline Processing**: The data pipeline, orchestrated using tools like AWS Lambda, processes incoming data, including ECG predictions, and routes it to the appropriate storage or database.

- **Dashboard Integration**:
  - **Data Dashboard Design**: user-friendly dashboard is designed to visualize ECG data and predictions. Dashboard is developed using HTML, CSS, Bootstrap and Bulma for the frontend and a simple Python-Flask server for the backend. All tasks are designed as microservices and serve this Flask backend server.
  - **API Consumption**: The dashboard consumes data from the API endpoints where ECG predictions and analysis results are made available.
  - **Real-Time Updates**: The dashboard is designed to provide real-time updates, enabling users to monitor ECG data and predictions as they are generated.

- **Data Security and Compliance**:
  - **Data Privacy**: Ensure compliance with data privacy regulations (e.g., HIPAA) by implementing robust data anonymization and access control measures. ECG data is stored with an anonymized identifier using the RedCap ID.
  - **Data Encryption**: Data is encrypted during transmission and storage to protect sensitive health information.

Refer to figure 1 for the whole system architecture.

## 3.1 Predictive Models and Evaluation

Our predictive models give the predictive analysis for heart conditions such as structural Heart Disease, LV Systolic Dysfunction (LVSD) and hypertrophic cardiomyopathy (HCM).

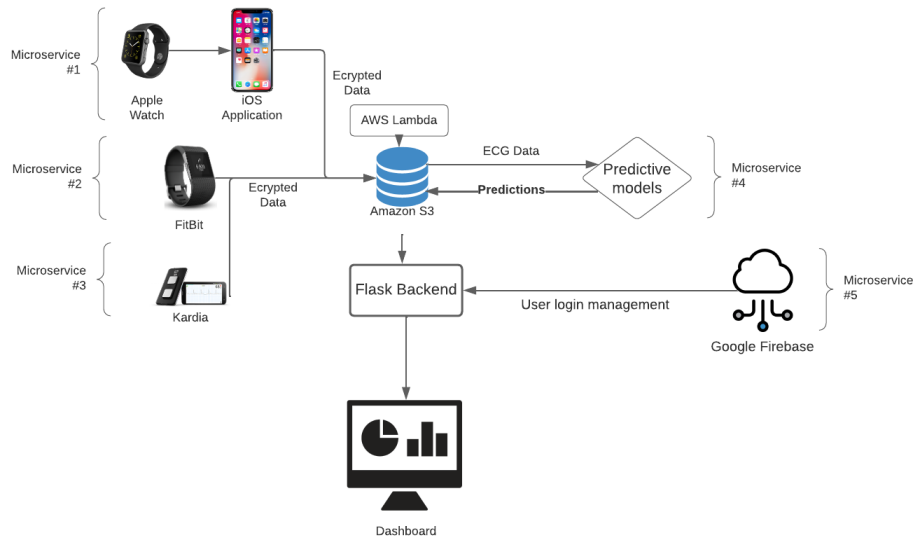

Figure 1: System Architecture

(Khunte **andothers** 2023) explains the model architecture for LVSD screening in detail. As for HCM, an XGBoost Classifier ensemble layer is used on top of a fine-tuned CNN architecture similar to the one used in (Khunte **andothers** 2023). The CNN architecture consisted of a (5000, 1, 1) input layer, corresponding to a 10-s, 500Hz, Lead I ECG, followed by seven two-dimensional convolutional layers, each of which were followed by a batch normalization layer31, ReLU activation layer, and a two-dimensional max-pooling layer. The output of the seventh convolutional layer was then taken as input into a fully connected network consisting of two dense layers, each of which were followed by a batch normalization layer, ReLU activation layer, and a dropout layer with a dropout rate of 0.532. The output layer was a dense layer with one class and a sigmoid activation function.

### 3.2 Dashboard and iOS Application

Representative images of the dashboard are presented in figures 4, 5, 6. Figure 3 demonstrates a screenshot of the iOS application developed to collect and submit an ECG associated with a Study ID.

### 3.3 Evaluation Approaches

To evaluate the performance of the architecture, five time tests were conducted to determine the average latency and turnaround time for the entire process of recording an ECG on two different devices (Apple Watch and Kardia). This evaluation has been broken down into parts depending on the functionality, to get a better analysis of the performance.

The backend of the dashboard checks for new Kardia and Apple Watch ECG data every 30 seconds. Any new data picked up, will be immediately pre-processed before running the ML models on them. This adds to the overall overhead time taken to get predictive analysis on the dashboard.

# 4 Results

## 4.1 Evaluation Against Source Data Compared One to One From a Single User's Watch and Kardia Acquisition

As described before, ECG data from the Apple watch can be acquired only through an iOS application. Refer to 3 for the prototype iOS application we have developed for this purpose. Apple watch records ECGs at 500Hz for a duration of 30 seconds. Once ECG is recorded on the apple watch, the application takes about 2 seconds to convert the recorded ECG into a numPy array and write it to the secure database. As for Kardia, the ECGs are recorded for 30 seconds at 100Hz. It takes about 3-4 seconds for our Kardia microservice to query the API to get the new ECG recording and add it to our database.

Once a new ECG has been added to the database, the AWS Lambda job picks it up and runs all the three predictive models on it. The results are stored in the database with relevant metadata. Our dashboard is dynamic and updated in near-real time. So as soon as the new predictions are available, it is updated with the ECG data, metadata and the corresponding predictive analysis.

Refer to the figure 2 for the ECG recording and the predictive analysis of a user from Kardia, as displayed in our dashboard.

## 4.2 Evaluation of time efficiency of the Platform

In the process of evaluation of the platform for time efficiency, we thoroughly assessed the performance of the platform in terms of the time it takes from the moment of data acquisition to the generation and reporting of results. This was done by running each of the Apple Watch and KardiaMobile scenarios 5 times, and calculating the mean time across these 5 trials. The mean duration observed in this process ranged from 33.0 to 35.7 seconds, after a standard 30-second data acquisition period. This efficiency enabled the entire procedure to be completed within a remarkably short timeframe, spanning from 63.0 to 65.7 seconds.

After the 30s of acquisition on KardiaMobile, it took a mean of 19.17 seconds for the backend to query the AliveCor API and pick up the new data, 11.49 seconds for running the models, and 2.35 seconds to write the results back to the S3 storage and results to be displayed on the dashboard, for a post-acquisition reporting time of 33 seconds, and a total time inclusive of acquisition to reporting of 63.0 seconds.

| Time Efficiency (all time values in seconds) | | | | | | |
|---|---|---|---|---|---|---|
| Device | Time to record ECG | Mean time for data to be uploaded to AWS S3 | Mean time for new ECG data to be picked up by the backend | Mean time to run predictive models on the data | Mean turnaround time for results to be displayed on the dashboard | Mean time for the entire process |
| Apple Watch | 30 | 0.7 | 19.17 | 13.51 | 2.35 | 65.73 |
| Kardia | 30 | 0 | 19.17 | 11.49 | 2.35 | 63.01 |

Table 1: Time efficiency of the entire architecture.

After the 30s of acquisition on an Apple Watch, it took a mean of 0.7 seconds to write the ECG XML file to an S3 bucket, 19.17 seconds for the backend to pick up the new data and convert it into the desired JSON format, 13.51 seconds for running the models, and 2.35 seconds to write the results back to the S3 storage and results to be displayed on the dashboard, for a post-acquisition reporting time of 35.73 seconds, and a total time inclusive of acquisition to reporting of 65.73 seconds.

Crucially, our evaluation revealed that there were no significant disparities in the time taken for acquisition and subsequent reporting when using two different commercially available devices, namely the Apple Watch and KardiaMobile. This uniformity across these devices highlights the robustness and consistency of the platform's performance.

## 5 Discussion

Through the convergence of wearable technology, data science, and predictive analytics, we have developed a unified ECG platform that harmonizes data collection, analysis, and visualization. Our prototype has been designed and developed with a special focus on the needs of end-users as well as the digital health community, creating a unified platform that can directly deploy and test at scale novel ECG-based screening, diagnostic and predictive tools. It offers real-time monitoring in the community and further promotes equitable access to cardiovascular care.

The whole architecture involves collecting ECG data from multiple wearable devices, employing data pre-processing techniques to ensure quality and consistency, leveraging advanced machine learning models for predictive analytics, and integrating the results into a robust data pipeline. All of the data is showcased on a dynamic and scalable dashboard with emphasis on data security and privacy.

The innovative aspect of this architecture lies in combining together the three key steps of collected data from various sources in near real-time, standardizing and deploying the AI models on Cloud, and retrieving the results from the models and displaying it to the end user almost instantly.

Such a platform offers a multitude of benefits, including but not limited to, early intervention, remote monitoring, cost efficiency, and data-driven insights.

# 6 Acknowledgements

**Funding:** Dr. Khera is supported by the National Heart, Lung, and Blood Institute of the NIH (under award K23HL153775) and the Doris Duke Charitable Foundation (under award 2022060).

**Disclosures:** Dr. Khera is an Associate Editor of JAMA. He also receives research support, through Yale, from Bristol-Myers Squibb and Novo Nordisk. He is a coinventor of U.S. Provisional Patent Applications 63/177,117, 63/428,569, 63/346,610, 63/484,426, and 63/508,315, all within the domain of machine learning in healthcare. He is also a co-founder of Evidence2Health and Ensight-AI, which are precision health platforms to improve cardiovascular diagnosis and evidence-based cardiovascular care.

# 7 Appendix

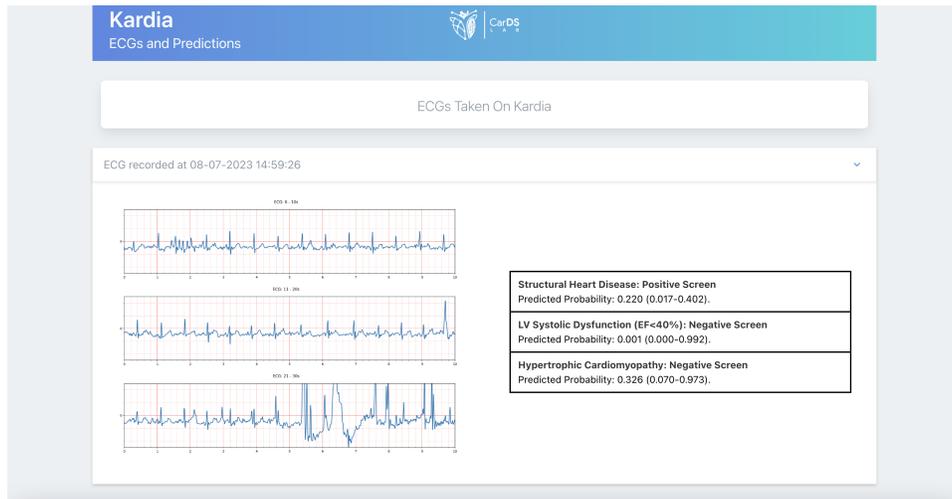

Figure 2: Predictive analysis of one of the ECG recordings from Kardia

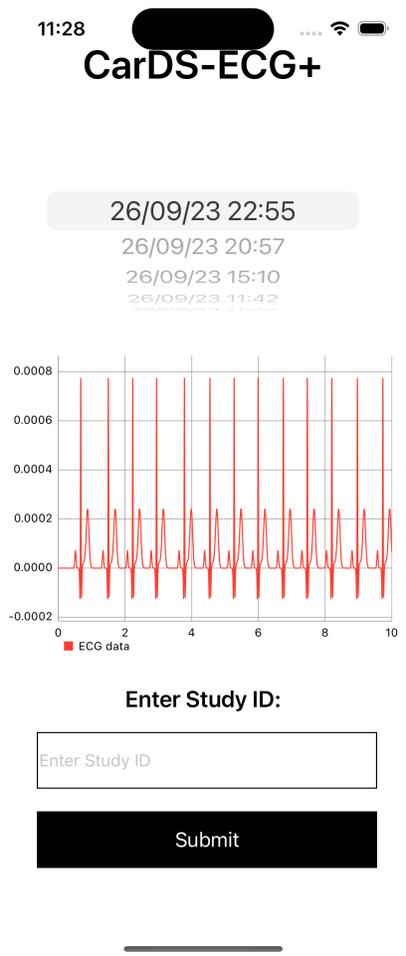

Figure 3: Prototype of the iOS application to collect ECG data from Apple Watches

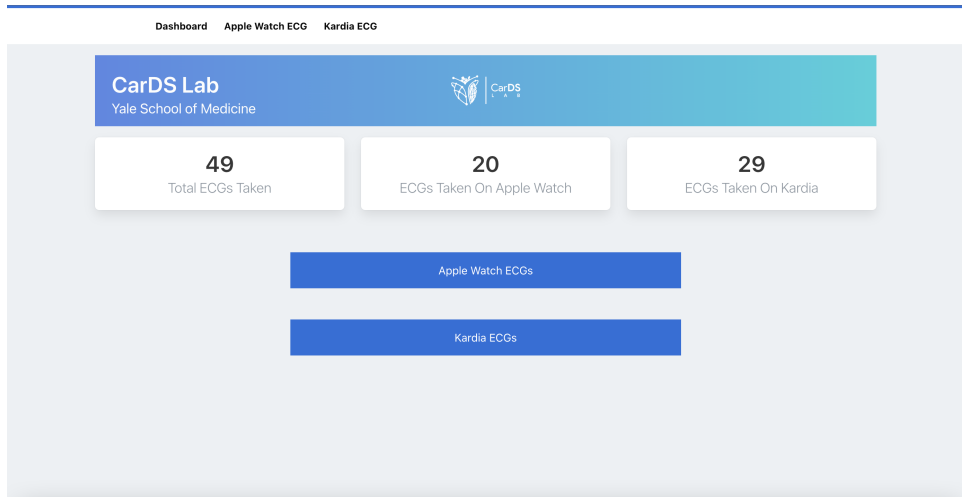

Figure 4: Home page of the prototype dashboard

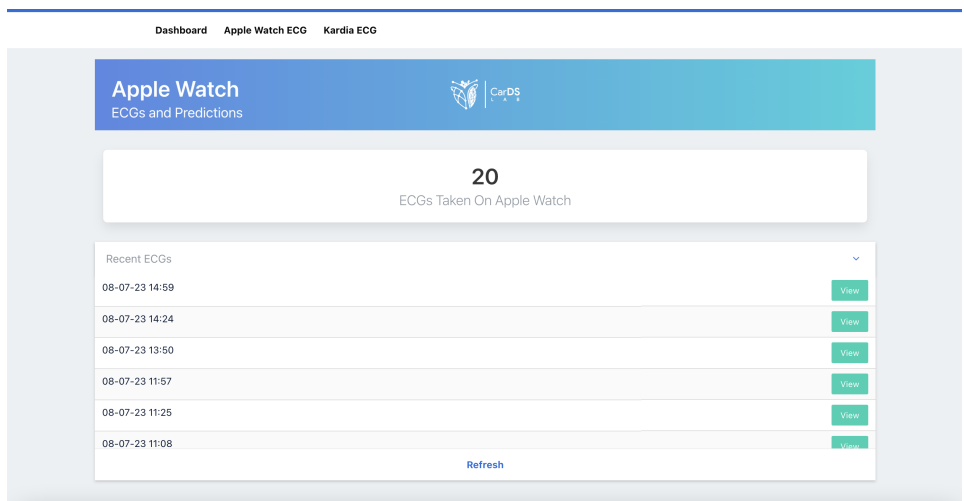

Figure 5: Apple watch ECG recordings

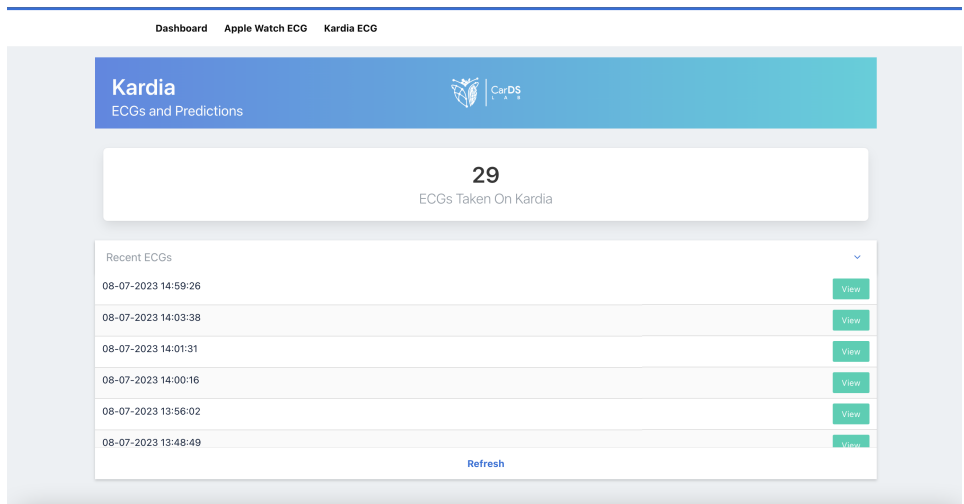

Figure 6: Kardia ECG recordings